\documentclass[10pt,twocolumn,letterpaper]{article}

\usepackage{cvpr}              

\usepackage[dvipsnames]{xcolor}

\definecolor{cvprblue}{rgb}{0.21,0.49,0.74}
\usepackage[pagebackref,breaklinks,colorlinks,citecolor=cvprblue]{hyperref}

\usepackage{multirow}
\usepackage{indentfirst}
\usepackage{booktabs}
\usepackage{caption}
\usepackage{tabularx}

\def\paperID{11630} 
\def\confName{CVPR}
\def\confYear{2024}

\title{Adaptive Multi-Modal Cross-Entropy Loss for Stereo Matching}

\author{Peng Xu \quad Zhiyu Xiang\thanks{Corresponding author.} \quad Chengyu Qiao \quad Jingyun Fu \quad Tianyu Pu\\
College of Information Science and Electronic Engineering,  Zhejiang University\\
{\tt\small \{xxxupeng, xiangzy, 3140104437, fujingyun, 3190105835\}@zju.edu.cn}
}

\begin{document}
\maketitle
\begin{abstract}

Despite the great success of deep learning in stereo matching, recovering accurate disparity maps is still challenging. Currently, L1 and cross-entropy are the two most widely used losses for stereo network training. Compared with the former, the latter usually performs better thanks to its probability modeling and direct supervision to the cost volume. However, how to accurately model the stereo ground-truth for cross-entropy loss remains largely under-explored. Existing works simply assume that the ground-truth distributions are uni-modal, which ignores the fact that most of the edge pixels can be multi-modal. In this paper, a novel adaptive multi-modal cross-entropy loss (ADL) is proposed to guide the networks to learn different distribution patterns for each pixel. Moreover, we optimize the disparity estimator to further alleviate the bleeding or misalignment artifacts in inference. Extensive experimental results show that our method is generic and can help classic stereo networks regain state-of-the-art performance. In particular, GANet with our method ranks $1^{st}$ on both the KITTI 2015 and 2012 benchmarks among the published methods. Meanwhile, excellent synthetic-to-realistic generalization performance can be achieved by simply replacing the traditional loss with ours. Code
is available at \textcolor{magenta}{https://github.com/xxxupeng/ADL}.
\end{abstract}    
\section{Introduction}


As a long-standing and active topic in computer vision, stereo matching plays an essential role in wide applications such as autonomous driving and virtual reality. While conventional methods suffer from poor reliability in tackling illumination change and weak texture, the learning-based stereo methods show their superiority in these complex scenes.


\begin{figure}[t]
  \centering

  \begin{subfigure}{0.90\linewidth}
    \includegraphics[width=\textwidth]{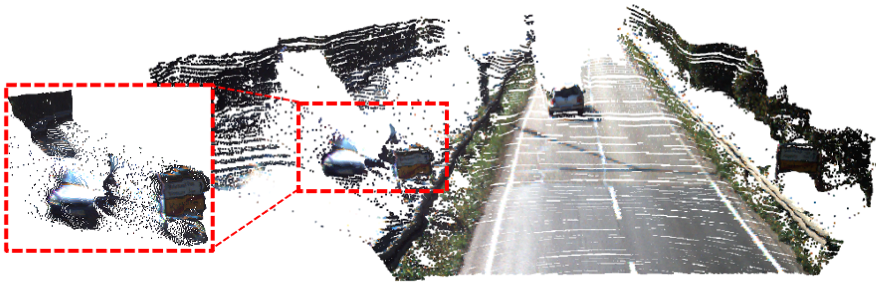}
    \caption{Over-smoothing artifacts from PSMNet~\cite{PSMNet}}
    \label{pcd 1}
  \end{subfigure}
  \begin{subfigure}{0.90\linewidth}
    \includegraphics[width=\textwidth]{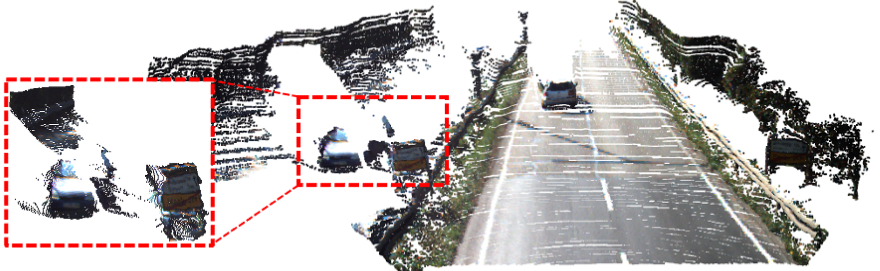}
    \caption{Misalignment artifacts from PSMNet + \cite{SMNet}}
    \label{pcd 2}
  \end{subfigure}
  \begin{subfigure}{0.90\linewidth}
    \includegraphics[width=\textwidth]{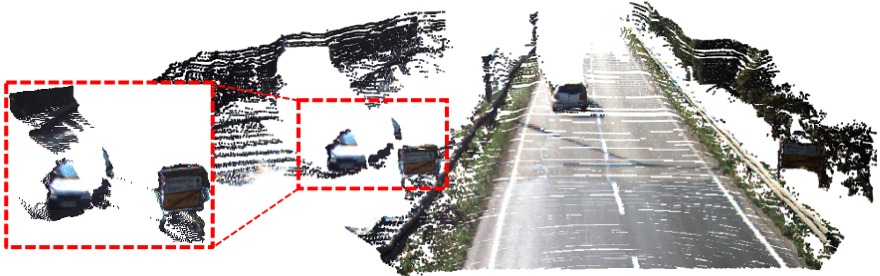}
    \caption{PSMNet + Ours}
    \label{pcd 3}
  \end{subfigure}

  \caption{\textbf{Comparison of the reconstructed point clouds.} Our method can alleviate the over-smoothing and misalignment artifacts, which is critical to the performance of downstream tasks.}
  \label{Point Cloud Comp.}
\end{figure}

Stereo matching is usually regarded as a regression task in deep learning~\cite{GCNet,PSMNet,SGNet,GwcNet,GANet}. In these works, L1 loss is employed for training, followed by the soft-argmax estimator~\cite{GCNet} to predict sub-pixel disparity. The main problem of L1 loss is that it lacks direct supervision of the cost volume and is thereby prone to overfitting~\cite{AcfNet}. Moreover, soft-argmax is based on the assumption that the output distributions are uni-modal and centered on the ground-truth~\cite{GCNet}, which is not always true especially for the edge pixels with ambiguous depths. As shown in~\cref{pcd 1}, soft-argmax on edge pixels suffers from severe over-smoothing problem, causing bleeding artifacts at the edge.

Another line of research treats stereo matching as a classification task, where the cross-entropy loss could be used. To guide the network to output uni-modal distributions, researchers model the ground-truth disparity with discrete Laplacian or Gaussian distributions~\cite{PDSNet,SMNet,AcfNet,LaC}. The single-modal disparity estimator (SME)~\cite{PDSNet,SMNet} is further employed to extract correct modals from the predicted distributions. The cross-entropy loss can directly supervise the learning of the cost volume, thereby achieving better results than the L1 loss. However, the enforcement of uni-modal pattern seems not that effective, as evidenced by the presence of misalignment artifacts in~\cref{pcd 2}.

Our work aims to explore a better modeling for the stereo ground-truth and improve the disparity estimator. Contrary to previous works that impose the uni-modal constraints on the cost volume, we believe that the edge pixels should naturally be modeled as the multi-modal distributions. During the image capture process, edge pixels collect lights from multiple objects at different depths, implying that the depth of edge pixels inherently carries ambiguity. Enforcing the network to learn the uni-modal pattern at all areas can be confusing and misleading, causing erroneous estimation on both edge and non-edge pixels. Therefore, a better probability model encoding the true patterns of each pixel is highly desirable.

In this paper, we propose adaptive multi-modal distribution model for pixels and integrate it into cross-entropy loss for network training. We apply disparity clustering within the local window of each pixel to obtain the desired number of the modals. Laplacian distribution is then employed for modeling each cluster. We further rely on the local structural information within the window to determine the relative weight of each modal, thereby finalizing the mixture of Laplacians for the cross-entropy loss. Additionally, we propose a dominant-modal disparity estimator (DME) to better tackle the difficulties brought by the multi-modal outputs from the network. Extensive experimental results on public datasets show that our method is generic and can help classic stereo networks regain state-of-the-art performance. The comparison results in~\cref{Point Cloud Comp.} exemplify the remarkable improvements of our method. Moreover, our method achieves excellent cross-domain generalization performance and exhibits higher robustness to sparser ground-truth.

Our contributions can be summarized as follows: 
\begin{itemize}
    \item We propose an adaptive multi-modal cross-entropy loss for training stereo networks. It can effectively guide the networks to learn clear distribution patterns and suppress outliers.
    \item We propose a dominant-modal disparity estimator that can obtain accurate results upon the multi-modal outputs.
    \item Extensive experiments show that our method is general and can help the classic stereo networks regain highly competitive performance. GANet~\cite{GANet} with our method ranks $1^{st}$ on both the KITTI 2015~\cite{KITTI2015} and the KITTI 2012~\cite{KITTI2012} benchmarks among all published methods.
    \item Networks with our method exhibit excellent generalization performance, surpassing existing methods that specialize in cross-domain generalization.
    \item Our method is robust to sparser supervision, revealing great potential to save the cost of producing dense ground-truth for network training.
\end{itemize}

\section{Related work}

\textbf{Deep stereo matching.}
DispNet~\cite{SceneFlow}, a model that constructs a correlation volume and directly regresses the disparity, is the first end-to-end deep stereo network. Later, GCNet~\cite{GCNet} proposes constructing the cost volume with concatenated features and employing 3D convolutions for cost aggregation. PSMNet~\cite{PSMNet}, the popular baseline for the following cost volume-based works~\cite{SMNet,SMDNet,AcfNet,ACVNet}, adds the spatial pyramid pooling~\cite{SPP} to the network and stacks multiple hourglass networks to improve the accuracy. GwcNet~\cite{GwcNet} further improves the cost volume by the group-wise correlation that provides more efficient measure of feature similarity. To reduce the computational complexity, GANet~\cite{GANet} proposes replacing the 3D convolutions with the aggregation layers guided by semi-global and local information. Following RAFT~\cite{raft}, another branch of work~\cite{igev,dlnr,eaistereo,RAFTStereo,CREStereo} relies on iterative refinement pipeline with ConvGRU~\cite{GRU} to achieve high disparity precision. Recently, IGEVStereo~\cite{igev} proposes combining a geometry encoding volume with the correlation feature in the iterative pipeline, achieving the state-of-the-art performance on KITTI 2015 benchmark~\cite{KITTI2015}.

\textbf{Loss function and disparity estimator.} Loss function and disparity estimator are crucial for stereo networks. The former supervises the learning process, and the latter finalizes the disparity from the distribution volume. In GCNet~\cite{GCNet}, regression-based L1 loss is adopted and the full-band weighted average operation (soft-argmax) is proposed to calculate the final disparity. Later, smooth L1 loss becomes the mainstream~\cite{PSMNet,GwcNet,PGNet,SGNet,ACVNet}. Different from the above works, PDSNet~\cite{PDSNet} and the following works~\cite{SMNet,AcfNet,LaC} uses the uni-modal cross-entropy loss to impose direct supervision to the distribution volume. No matter what the loss function is, the multi-modal outputs caused by the matching ambiguity are unavoidable. Soft-argmax on these multi-modal outptus leads to over-smoothing artifacts on the edge pixels. To solve this problem, SME~\cite{PDSNet,SMNet} selects the modal with the maximum probability and only estimates the final disparity on it. CDN~\cite{CDN} determines the integer part of the disparity from the modal with maximum probability and further estimates the offsets by a small network. SMDNet~\cite{SMDNet} feeds the distribution volume to MLPs~\cite{mlp} to parameterize the network outputs as the mixture of two Laplacians, and chooses the modal with higher peak as the final result. Beside these post-processing methods,~\cite{Non-parametric} notices the multi-modal nature of the ground-truth when supervising the coarse-level cost volume in their multi-view stereo study. Contrary to the existing works that impose uni-modal distribution for each pixel, our method models ground-truth as adaptive multi-modal distributions and encourages multi-modal outputs on edge pixels. Different from~\cite{Non-parametric} that introduces multi-hot cross-entropy loss for coarse-level patch-sized pixels but still employs L1 loss for the fine-level outputs, our method directly sets up adaptive multi-modal loss for the fine-level pixels, providing more direct and effective supervision to the network. We also optimize the disparity estimator to better tackle the multi-modal outputs from the distribution volume.

\textbf{Cross-domain generalization.}
As another important issue for deep learning-based stereo matching, capability of cross-domain generalization has been extensively studied. DSMNet~\cite{DSMNet} proposes a novel domain normalization layer combined with a learnable non-local graph-based filtering layer to reduce the domain shifts. CFNet~\cite{CFNet} builds a cascade and fused cost volume representation to learn domain-invariant geometric scene information. ITSA~\cite{ITSA} refers to the information bottleneck principle~\cite{bottleneck1} to minimize the sensitivity of the feature representations to the domain variation. GraftNet~\cite{GraftNet} embeds a feature extractor pre-trained on large-scale datasets into the stereo matching network to extract broad-spectrum features. Without adding any additional learnable modules, we achieve outstanding generalization performance by simply changing the training loss.

\section{Method}
\subsection{Fundamentals and problem statement}
\label{Problem statement}


Given a calibrated stereo image pair, stereo matching aims to find the corresponding pixel in the right image for each pixel in the left image. The cost volume-based stereo networks follow the common pipeline~\cite{GCNet}. First, features of the left and right images are extracted by a weight-sharing 2D CNN module respectively. Then a 4D cost volume is constructed upon the two obtained feature blocks. The cost aggregation module takes this 4D volume as input and outputs a distribution volume with size $D \times H \times W$, where $D$ is the maximum range of disparity search, $H$ and $W$ are the height and width of the input image, respectively. Softmax operator is then applied along the disparity dimension to normalize the probability distribution $p(\cdot)$ for each pixel. Finally, the resulting disparity $\hat d$ is estimated by the full-band weighted average operation, which is also called soft-argmax:

\begin{equation} 
\hat{d}=\sum_{d=0}^{D-1}d \cdot p(d) \label{full band}
\end{equation}

To train the stereo network, regression-based smooth L1 loss can be employed with:

\begin{equation}
\mathcal{L}_{reg}(\hat d,d_{gt} ) = 
\begin{cases}
0.5(\hat d - d_{gt} )^2, & if\ |\hat d - d_{gt}|<1, \\
|\hat d -d_{gt}|-0.5, & otherwise. \\
\end{cases} \label{smooth l1}
\end{equation}

\noindent where $d_{gt}$ is the ground-truth disparity.

\begin{figure}[t]
    \centering
    \includegraphics[width=0.8\linewidth]{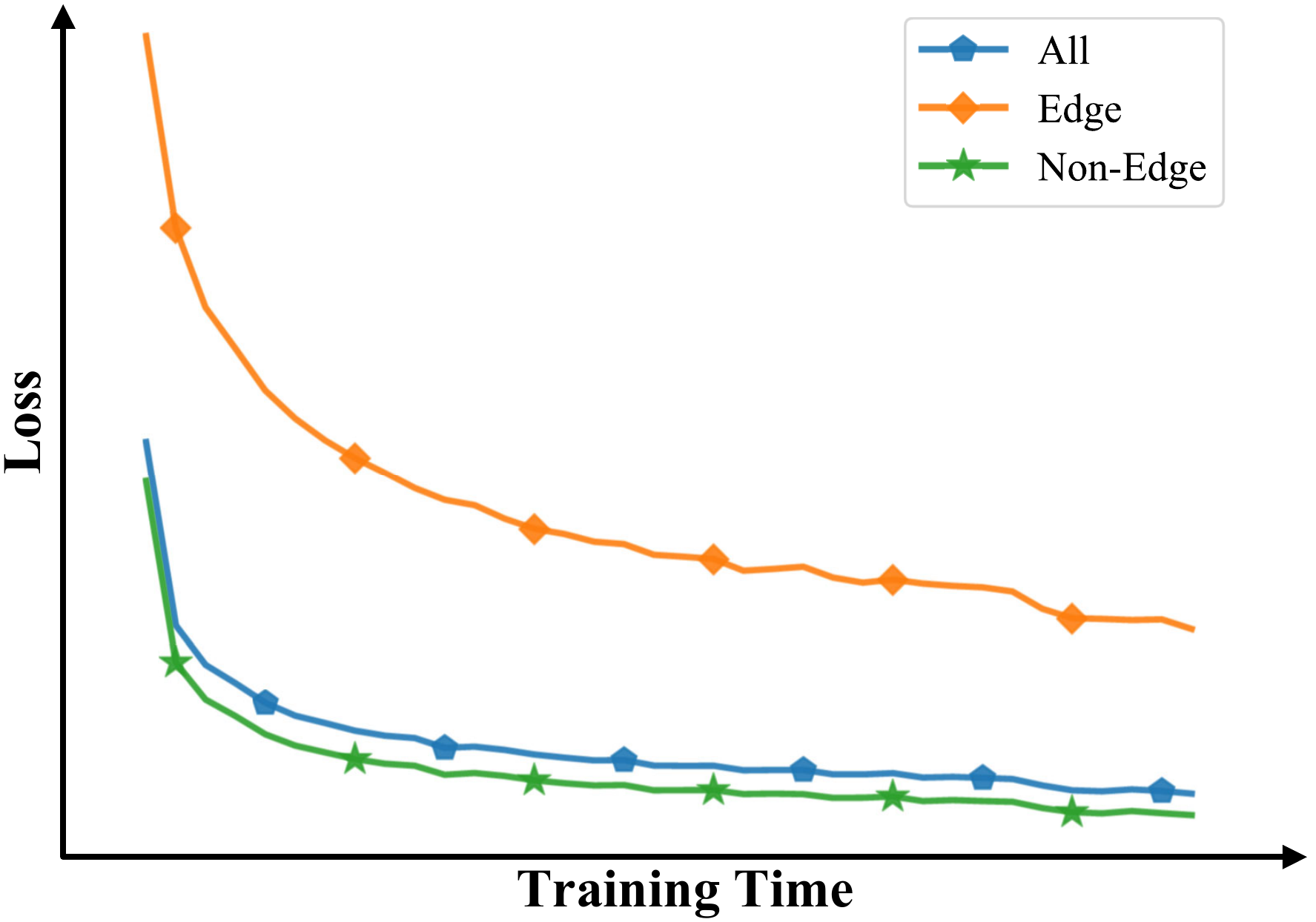}
    \caption{\textbf{Training trends} of the uni-modal cross-entropy loss on SceneFlow dataset.}
    \label{loss}
\end{figure}

In this pipeline, the distribution volume is indirectly supervised by the smooth L1 loss, which hinders the final performance~\cite{AcfNet}. By treating the stereo matching as a classification task, cross-entropy loss provides direct supervision on the distribution volume, as:

\begin{equation}
\mathcal{L}_{ce}(p,p_{gt})= -\sum_{d=0}^{D-1}p_{gt}(d) \cdot {\rm log}\ p(d)\label{cross entropy}
\end{equation}

\begin{figure*}[t]
\centering
\includegraphics[width=0.95\linewidth]{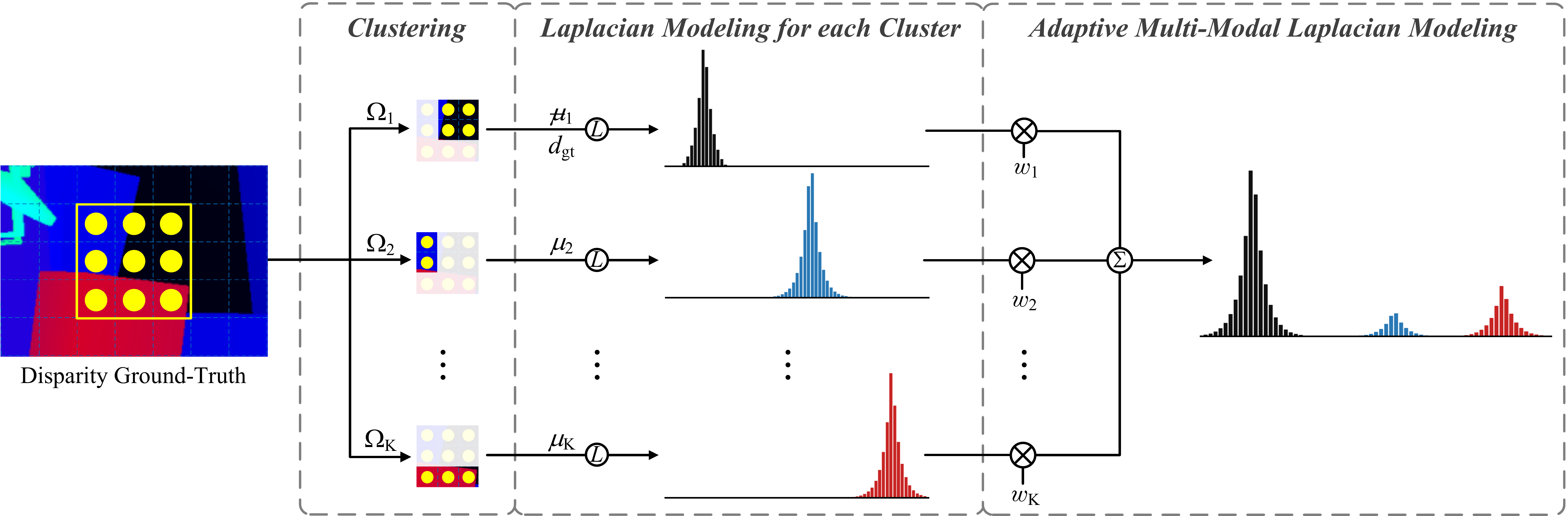}
\caption{\textbf{Illustration of our adaptive multi-modal modeling} for cross-entropy loss. Given the pixel for modeling, the disparities within a pre-defined window are divided into $K$ clusters $\{\Omega_1,\Omega_2,...,\Omega_K\}$, and the mean $\mu_k$ for each cluster is calculated to form a \textit{uni-modal} Laplacian distribution. The final adaptive \textit{multi-modal} distribution is generated by the weighted summation of the Laplacian distributions, with the weight $w_k$ determined by $|\Omega_k|$.}
\vspace{-5pt}
\label{generation process}
\end{figure*}

The new problem is that the ground-truth distribution $p_{gt}(\cdot)$ in~\cref{cross entropy} is unavailable. Existing works~\cite{PDSNet,SMNet, AcfNet,LaC} simply model $p_{gt}(\cdot)$ as the uni-modal Laplacian or Gaussian distribution centered on $d_{gt}$. However, these simple models seem unable to impose sufficient supervision for different image regions, especially the edge. As shown in~\cref{loss}, the training loss of edge pixels remain much larger than that of the non-edge pixels, indicating the difficulty of learning in these areas. Further statistics (details shown later in~\cref{modal number}) on the resulting output distribution volume shows that over half of the edge and part of the non-edge pixels are actually assigned more than one modal, which conflicts with the uni-modal assumption of the pseudo-groundtruth. These undesired multi-modal outputs directly lead to the misalignment artifacts on object edges and outliers on non-edge areas, as shown in~\cref{pcd 2}. Therefore, we believe the root of the problem lies in the inappropriate uni-modal modeling of the ground-truth in all areas. In fact, edge pixels aggregate photometric information from multiple objects at different depths, implying that the intensities of edge pixels are inherently ambiguous. Imposing a uni-modal distribution pattern across the entire image will not only cause learning difficulties for edge pixels, but also confuse the learning for non-edge pixels.



\subsection{Adaptive multi-modal probability modeling}
\label{Adaptive Multi-Modal Cross-Entropy Loss}

Inspired by the observation in the previous section, we are dedicated to exploring a better probability modeling of ground-truth for the cross-entropy loss. We believe that the probability distributions of edge pixels should be composed of multiple modals, with each corresponding to a specific depth/disparity. To this end, an adaptive multi-modal ground-truth modeling method is proposed. Our idea is to generate a separate Laplacian distribution for each potential depth on the edge pixels and then fuse them together to construct a mixture of Laplacians. We refer to the neighborhood of each pixel to accomplish the task, as illustrated in~\cref{generation process}.


For each pixel labeled with ground-truth disparity, we consider a $m \times n$ local window centered on it. The entire set of disparity values within the window is then divided into $K (K\geq1)$ disjoint subsets $\{\Omega_1,\Omega_2,...,\Omega_K\}$ by the DBScan clustering algorithm~\cite{dbscan}, with each cluster corresponding to a different potential depth. In DBScan, the distance threshold $\epsilon$ and density threshold $minPts$ are set manually to adjust the resulting number of clusters. This clustering method offers the following advantages: (1) there is no need to pre-define the number of clusters; (2) $K=1$ can be regarded as an indicator of non-edges; (3) it is robust for slanted planes with continuous but varying depths.

The ground-truth distribution of each pixel can then be modeled as the mixture of Laplacians:

\begin{equation}
    \begin{aligned}
    p_{gt}(d) &= \sum_{k=1}^{K}w_k \cdot {\rm Laplacian}_{\mu_k,b_k}(d) \\
     &= \sum_{k=1}^{K}w_k \cdot \frac{e^\frac{-|d-\mu_k|}{b_k}}{\sum_{d_i=0}^{D-1}e^\frac{-|d_i-\mu_k|}{b_k}}
     \end{aligned}
    \label{multi-modal modeling}
\end{equation}

\noindent where the Laplacians are discretized and normalized over the disparity candidates $d \in \{0,1,...,D-1\}$, and $\mu_k$, $b_k$, and $w_k$ are the mean, scale, and weight parameters for the $k^{th}$ Laplacian distribution, respectively. $\mu_k$ is set to the mean value of the disparities within the cluster $\Omega_k$. Defining that $\Omega_1$ contains the central pixel to be modeled, $\mu_1$ is replaced by the central pixel's ground-truth to ensure the accuracy of the supervision. The weight $w_k$ is designed to adjust the relative proportions of the obtained multiple modals, and can be assigned based on the local structure within the window. We take the cardinality of $\Omega_k$ as an indicator of the local structure, \eg, a smaller $|\Omega_k|$ corresponds to a thinner structure, which should have smaller weight accordingly. Finally, $w_k$ is defined as:

\begin{equation}
    w_k = 
    \begin{cases}
        \alpha +  (|\Omega_k|-1) \cdot \frac{ 1-\alpha}{mn-1}, & k=1\\
        
        |\Omega_k| \cdot \frac{1-\alpha}{mn-1}, & k \neq 1
    \end{cases}
    \label{weight parameter}
\end{equation}

\noindent where $\alpha$ is a fixed weight for the central pixel. We set $\alpha \geq 0.5$ to ensure the dominance of the ground-truth modal. The rest $(1-\alpha)$ weights are equally distributed to the rest $(mn-1)$ neighboring pixels. For datasets with sparse ground-truth like KITTI~\cite{KITTI2012,KITTI2015}, only valid disparities within the local window are counted and $mn$ in~\cref{weight parameter} is replaced with $\sum_{k=1}^K|\Omega_k|$. For non-edge pixels which have only one cluster within the window, $w_1$ is equal to one, and~\cref{multi-modal modeling} degenerates into a uni-modal Laplacian distribution.

\subsection{Dominant-modal disparity estimator}

Stereo networks trained by cross-entropy loss often yield more multi-modal outputs than the L1 loss, thereby a better disparity estimator is highly desired. SME~\cite{SMNet} alleviates the over-smoothing problem by aggregating disparities only within the “most likely” modal, but still suffers from the misalignment artifacts caused by the erroneous modal selection. In contrast to the uni-modal loss, our new loss encourages the network to generate more multi-modal patterns at the edge. Consequently, an enhanced disparity estimator becomes imperative to address the complexities introduced by our method. 

\definecolor{plt_green}{rgb}{0.172549, 0.62745, 0.172549}
\definecolor{plt_blue}{rgb}{0.12156863, 0.46666667, 0.705882}
\definecolor{plt_light_blue}{rgb}{0.55686, 0.72941, 0.85098}

\begin{figure}[t]
    \centering

    \includegraphics[width=0.95\linewidth]{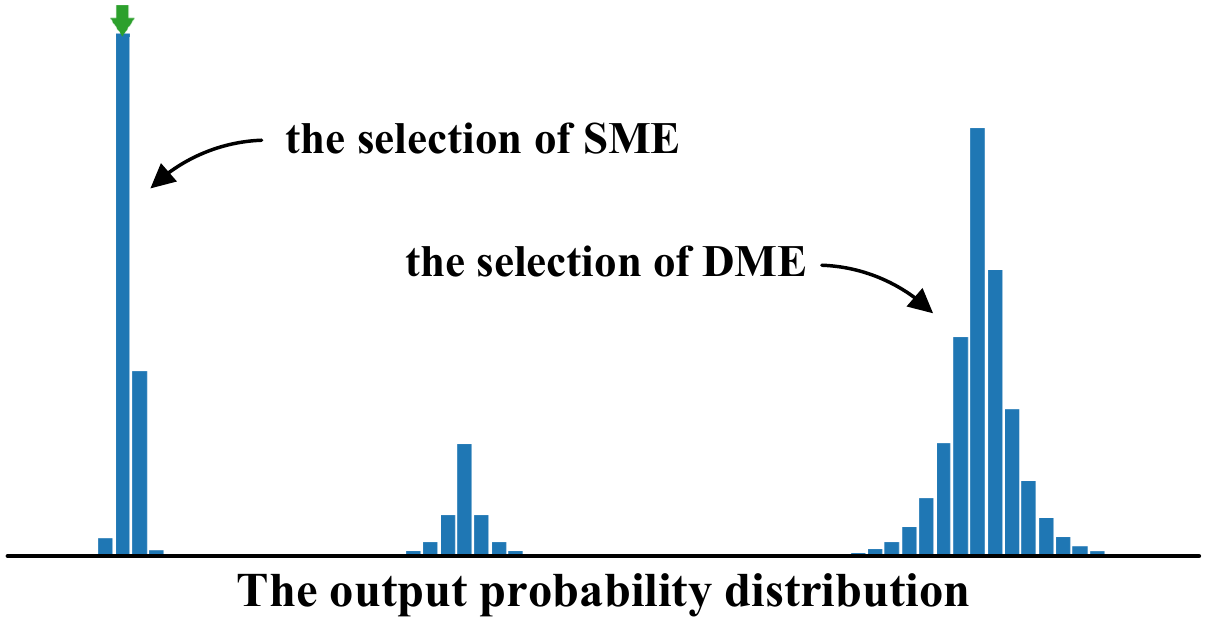}
    \caption{\textbf{Illustration of modal selection strategy} during inference. SME~\cite{SMNet} prefers the modal with maximum probability candidate (aimed by the \textcolor{plt_green}{green arrow}). Our proposed DME prefers the one with maximum cumulative probability.}
    \label{disparity estimator}

\end{figure}

SME first locates the disparity candidate with the maximum probability density, and then traverses left and right respectively until the probability stops decreasing, thereby determining the range of the dominant modal $[d_l,d_r]$ for disparity estimation. However, this pixel-level winner-take-all strategy is sensitive to noises that can produce sharp and narrow modals in the output distribution.

We propose our DME to solve this problem, as illustrated in~\cref{disparity estimator}. Specifically, we split each modal from the multi-modal output and calculate their cumulative probability separately. Each modal corresponds to a potential matching object with a specific depth, and the cumulative probability of the modal reflects the matching possibility of that object. Therefore, we adopt an object-level winner-take-all strategy, \ie, selecting the modal with the maximum cumulative probability as the dominant modal. The selected modal is then normalized as:

\begin{equation}
\overline p(d) = 
\begin{cases}
\frac{p(d)}{\sum_{d_i=d_l}^{d_r}p(d_i)}, & if\ d_l \leq d \leq d_r, \\
0, & otherwise. \\
\end{cases} \label{distribution norm.}
\end{equation}

Finally,~\cref{full band} is used to estimate the disparity by substituting $\overline p(d)$ for $p(d)$.

\begin{table*}[t] \footnotesize
    \centering
    \setlength{\tabcolsep}{8pt}
    \begin{tabular}{l c c c c c c c}
        \toprule
        \multirow{2}{*}{Method} & \multicolumn{2}{c}{Cross-Entropy Loss} & \multicolumn{2}{c}{Disparity Estimator} & \multirow{2}{*}{EPE} & \multirow{2}{*}{\textgreater1px} & \multirow{2}{*}{\textgreater3px}  \\
        \cmidrule(lr){2-3} \cmidrule(lr){4-5}
         & Uni-Modal (UM) & Multi-Modal (MM) & SME & DME \\
         \midrule
         PSMNet~\cite{PSMNet} & \multicolumn{4}{c}{}
         & 0.97 & 10.51 & 4.03 \\
         \midrule
         PSMNet + UM + SME~\cite{SMNet} & \checkmark & & \checkmark & & 0.84 & 6.65 & 2.85\\
         PSMNet + UM + DME & \checkmark & & & \checkmark & 0.82 & 6.61 & 2.82 \\
         PSMNet + MM + SME & & \checkmark & \checkmark & & 0.80 & 6.31 & 2.72\\
         PSMNet + MM + DME (Ours) & & \checkmark & & \checkmark & \textbf{0.78} & \textbf{6.30} & \textbf{2.71}\\
         \bottomrule
         
    \end{tabular}
    \caption{\textbf{Ablation study} of the loss function in training and the disparity estimator in inference on SceneFlow.}
    \label{ablation study loss and estimator}
\end{table*}

\begin{table}[t]  \footnotesize
    \centering
    \setlength{\tabcolsep}{10pt}
    \begin{tabular}{c c c c c c}
         \toprule
          \multicolumn{2}{c}{Local Window} & EPE & \textgreater1px & \textgreater3px \\
         \midrule
         \multirow{3}{*}{Shape} & $3\times3$ & 0.83 & 6.67 & 2.86\\
         & $1\times9$ & \textbf{0.78} & \textbf{6.30} & \textbf{2.71} \\
         & $9\times1$ & 0.84 & 6.65 & 2.87 \\
         \midrule
         \multirow{5}{*}{Size} & $1\times3$ & 0.82 & 6.76 & 2.85\\
         & $1\times5$ & 0.81 & 6.58 & 2.79\\
         & $1\times7$ & 0.81 & 6.62 & 2.83\\
         & $1\times9$ & \textbf{0.78} & \textbf{6.30} & \textbf{2.71} \\
         & $1\times11$ & 0.84 & 6.72 & 2.91\\         
         \bottomrule
    \end{tabular}
    \caption{\textbf{Ablation study} of window shape and size on SceneFlow.}
    \label{window shape tab}
\end{table}

\section{Experiments}
\subsection{Datasets and evaluation metrics}

We evaluate our method on five popular stereo datasets. SceneFlow~\cite{SceneFlow} is a large synthetic dataset containing 35454 image pairs for training and 4370 for testing. KITTI 2012~\cite{KITTI2012} and KITTI 2015~\cite{KITTI2015} are the two  real outdoor datasets, each containing hundreds of images collected from driving scenes. Middlebury~\cite{Middlebury} and ETH3D~\cite{ETH3D} are also real-world datasets, with a few dozen of image pairs acquired in indoor or outdoor scenes. We only use the training sets of Middlebury and ETH3D to additionally validate the cross-domain generalization performance of our method.

As usual, EPE (End-Point-Error) and $k$px (the percentage of outliers with an absolute error greater than $k$ pixels) are employed to evaluate the networks' performance. For KITTI 2015, D1 metric (the percentage of disparity outliers) is reported.

\subsection{Implementation details}
We separately apply our method to three classic cost volume-based stereo networks, namely, PSMNet \cite{PSMNet}, GwcNet \cite{GwcNet}, and GANet \cite{GANet}. We implement all networks in PyTorch and use Adam with $\beta_1 = 0.9$ and $\beta_2 = 0.999$ as the optimizer. We train the networks from scratch using two NVIDIA 3090 GPUs. When training on SceneFlow, the learning rate is set to  $1\times 10^{-3}$  for the first 30 epochs and then reduced to $1\times 10^{-4}$ for the rest 15 epochs. On KITTI 2012 and 2015, we fine-tune the SceneFlow pre-trained networks for 600 epochs with the learning rate of $1\times 10^{-3}$. $b_k$ in~\cref{multi-modal modeling} and $\alpha$ in~\cref{weight parameter} are both set to 0.8 after parameter tuning. The distance threshold $\epsilon$ and the density threshold $minPts$ in DBScan~\cite{dbscan} are set to 3 and 1, respectively.


\subsection{Ablation study}
\label{ablation study}

We perform ablations on the SceneFlow dataset. The original PSMNet~\cite{PSMNet} trained with smooth L1 loss is taken as the baseline for comparison. To ensure fairness, all models are trained from scratch  for 15 epochs on the training set and then validated on the test set.

\textbf{Loss function.}
As shown in~\cref{ablation study loss and estimator}, compared with the baseline, PSMNet with our multi-modal cross-entropy loss boosts performance drastically, with EPE by 19.59\%, 1px error by 40.06\%, and 3px error by 32.75\%. Our method also outperforms the uni-modal cross-entropy loss~\cite{SMNet} on all metrics, demonstrating its superiority in effectively guiding the network to learn the explicit distribution patterns.

\textbf{Disparity estimator.}
We validate our disparity estimator by comparing with SME~\cite{SMNet}. As shown in~\cref{ablation study loss and estimator}, our DME consistently outperforms SME for the networks trained with either uni- or multi-modal losses, which proves its effectiveness in selecting the correct modals from multi-modal outputs.

\textbf{Window shape and size.}
The disparities within the window are used to construct the ground-truth distributions. We ablate to determine the optimal window shape and size.
~\cref{window shape tab} shows that local windows with horizontal 1D shape usually perform better than others and the $1 \times 9$ window achieves the best. This can be attributed to the nature of stereo matching, \ie, a 1D matching task along the horizontal direction.



\subsection{Analysis of the output distribution patterns}


\begin{table}[t] \footnotesize
\centering
\begin{tabular}{llcccc}
\toprule
\multirow{2}{*}{Method} & \multirow{2}{*}{Region} & \multicolumn{3}{c}{The number of modals} & \multirow{2}{*}{Outliers}\\
\cmidrule(lr){3-5}
& & 1 & 2 & $\geq$3\\
\midrule
\multirow{3}{*}{\shortstack[l]{PSMNet\\\cite{PSMNet}}}
& All & 98.50 & 1.00 & 0.50 & 4.03\\
& Edge & 87.79 & 10.30 & 1.91 & 25.59\\
& Non-Edge & 98.92 & 0.64 & 0.44 & 3.17\\
\midrule
\multirow{3}{*}{\shortstack[l]{+UM\\+SME\\\cite{SMNet}}}
& All & 94.67 & 4.60 & 0.73 & 2.85\\
& Edge & 40.16 & 52.70 & 7.14 & 19.39\\
& Non-Edge & 96.78 & 2.77 & 0.45 & 2.20\\
\midrule
\multirow{3}{*}{\shortstack[l]{+MM\\+DME\\(Ours)}}
& All & 94.92 & 4.35 & 0.73 & 2.71\\
& Edge & 35.35 & 57.28 & 7.37 & 18.97\\
& Non-Edge & 97.23 & 2.33 & 0.44 & 2.07\\
\bottomrule
\end{tabular}
\caption{\textbf{Statistics of pixels \wrt the number of modals and their corresponding outliers} (\textgreater3px) for PSMNet variants on SceneFlow. Modals with the peak density lower than 1\% are not included in the statistics.}
\label{modal number}
\end{table}


\begin{figure}[t]
    \centering
    \includegraphics[width=1.0\linewidth]{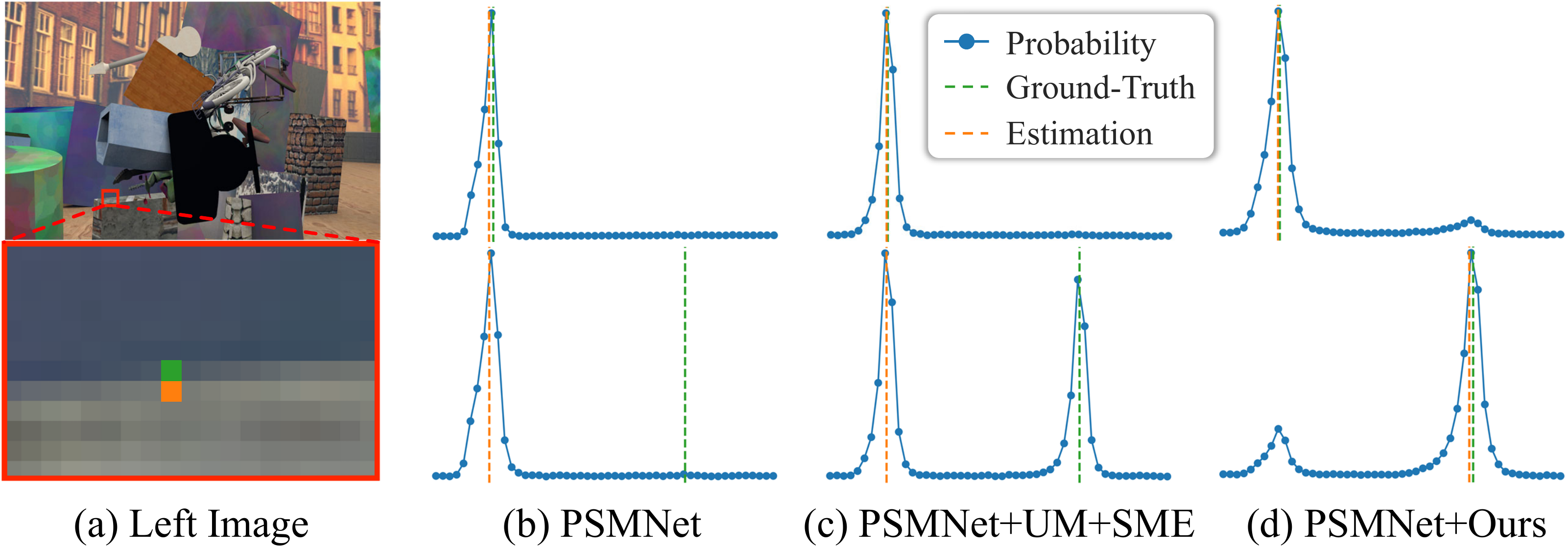}
    \caption{Visualization of output distributions at the edge. Top row: \textcolor[HTML]{2CA02C}{background} pixel,  bottom row: \textcolor[HTML]{FF7F0E}{foreground} pixel.}
    \label{probability}
\end{figure}

\begin{table}[t] \footnotesize
\centering
\setlength{\tabcolsep}{10pt}
\begin{tabular}{l c c c} 
\toprule
Method & EPE &  \textgreater1px & \textgreater3px  \\
\midrule
PSMNet~\cite{PSMNet} & 1.09 & 12.1 & 4.56 \\
AcfNet~\cite{AcfNet} & 0.87 & -- & 4.31 \\
GANet~\cite{GANet} & 0.78 & 8.70 & -- \\
GwcNet~\cite{GwcNet} & 0.77 & 8.00 & 3.30 \\
PSMNet +~\cite{SMNet} & 0.77 & -- & 2.21 \\
IGEVStereo~\cite{igev} & 0.47 & -- & 2.47\\
ACVNet~\cite{ACVNet} & \textbf{0.46} & 4.89 & 1.98 \\
\midrule
PSMNet + Ours & 0.64 & 5.14 & 2.19 \\
GwcNet + Ours & 0.62 & 5.07 & 2.16 \\
GANet + Ours & 0.50 & \textbf{4.25} & \textbf{1.81} \\
\bottomrule
\end{tabular}
\caption{\textbf{Quantitative results on SceneFlow test set.}}
\label{SceneFlow Performance Table}
\end{table}

\begin{figure}[t]
\centering
\includegraphics[width=0.95\linewidth]{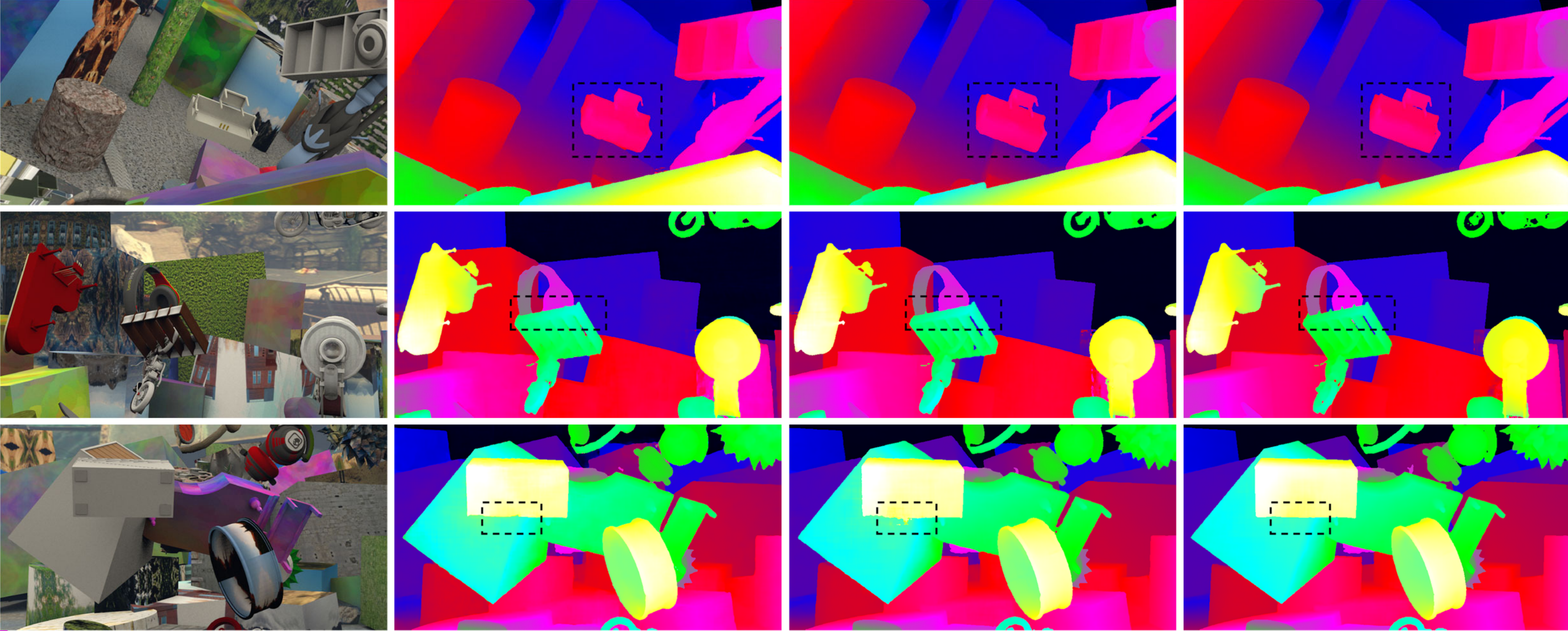}
\caption{\textbf{Qualitative comparison on SceneFlow.} From left to right: input images, disparity results from PSMNet, PSMNet+\cite{SMNet}, and PSMNet+Ours. Our method can recover more accurate object structure and reduce undesired defects at the edge.}
\vspace{-10pt}
\label{SceneFlow Qualitative Result}
\end{figure}

\begin{table*}[t] \footnotesize
\centering
\setlength{\tabcolsep}{8pt}
\begin{tabular}{l c c cc c c c cc c}
\toprule
\multirow{3}{*}{Method} & \multicolumn{6}{c}{KITTI 2015} & \multicolumn{4}{c}{KITTI 2012}\\
& \multicolumn{3}{c}{All} & \multicolumn{3}{c}{Noc}  & \multicolumn{2}{c}{\textgreater2px}  & \multicolumn{2}{c}{\textgreater3px}\\
\cmidrule(lr){2-4} \cmidrule(lr){5-7} \cmidrule(lr){8-9} \cmidrule(lr){10-11} 
& D1-bg & D1-fg & \textcolor{red}{D1-all} & D1-bg & D1-fg & D1-all & Out-Noc & Out-All & Out-Noc & Out-All\\
\midrule
PDSNet \cite{PDSNet} & 2.29 & 4.05 & 2.58 & 2.09 & 3.68 & 2.36 & 3.82 & 4.65 & 1.92 & 2.53 \\
PSMNet~\cite{PSMNet} & 1.86 & 4.62 & 2.32 & 1.71 & 4.31 & 2.14 & 2.44 & 3.01 & 1.49 & 1.89 \\
PSMNet +~\cite{SMNet} & 1.54 & 4.33 & 2.14 & 1.70 & 3.90 & 1.93 & 2.17 & 2.81 & 1.35 & 1.81\\
GwcNet~\cite{GwcNet} & 1.74 & 3.93 & 2.11 & 1.61 & 3.49 & 1.92 & 2.16 & 2.71 & 1.32 & 1.70 \\
PSMNet + SMDNet~\cite{SMDNet} & 1.69 & 4.01 & 2.08 & 1.54 & 3.70 & 1.89 & -- & -- & -- & -- \\
CDN~\cite{CDN} & 1.66 & 3.20 & 1.92 & 1.50 & 2.79 & 1.72 & -- & -- & -- & -- \\
AcfNet~\cite{AcfNet} & 1.51 & 3.80 & 1.89 & 1.43 & 3.25 & 1.73 & 1.83 & 2.35 & 1.17 & 1.54 \\
PSMNet +~\cite{Non-parametric} * & 1.56 & 3.49 & 1.88 & 1.42 & 3.29 & 1.73 & -- & -- & -- & -- \\
GANet~\cite{GANet} & 1.48 & 3.46 & 1.81 & 1.34 & 3.11 & 1.63 & 1.89 & 2.50 & 1.19 & 1.60 \\
GANet + LaC~\cite{LaC} & 1.44 & 2.83 & 1.67 & 1.26 & 2.64 & 1.49 & 1.72 & 2.26 & 1.05 & 1.42\\
ACVNet~\cite{ACVNet} & \textbf{1.37} & 3.07 & 1.65 & 1.26 & 2.84 & 1.52 & 1.83 & 2.34 & 1.13 & 1.47\\
LEAStereo~\cite{LEAStereo} & 1.40 & 2.91 & 1.65 & 1.29 & 2.65 & 1.51 & 1.90 & 2.39 & 1.13 & 1.45\\
IGEVStereo~\cite{igev} & 1.38 & 2.67 & 1.59 & 1.27 & 2.62 & 1.49 & 1.71 & 2.17 & 1.12 & 1.44 \\
CroCoStereo~\cite{weinzaepfel2023croco} & 1.38 & 2.65 & 1.59 & 1.30 & 2.56 & 1.51 & -- & -- & -- & -- \\
\midrule
PSMNet + Ours & 1.44 & 3.25 & 1.74 & 1.30 & 3.04 & 1.59 & 1.80 & 2.32 & 1.14 & 1.50 \\
GwcNet + Ours & 1.42 & 3.01 & 1.68 & 1.30 & 2.76 & 1.54 & 1.65 & 2.17 & 1.05 & 1.42\\
GANet + Ours & 1.38 & \textbf{2.38} & \textbf{1.55} & \textbf{1.24} & \textbf{2.18} & \textbf{1.40} & \textbf{1.52} & \textbf{2.01} & \textbf{0.98} & \textbf{1.29} \\
\bottomrule
\end{tabular}
\caption{\textbf{Quantitative results on KITTI 2015 and 2012 Benchmarks.} * retrained network.}\label{KITTI performance}

\end{table*}

\begin{figure*}
    \centering
    \includegraphics[width=0.95\linewidth]{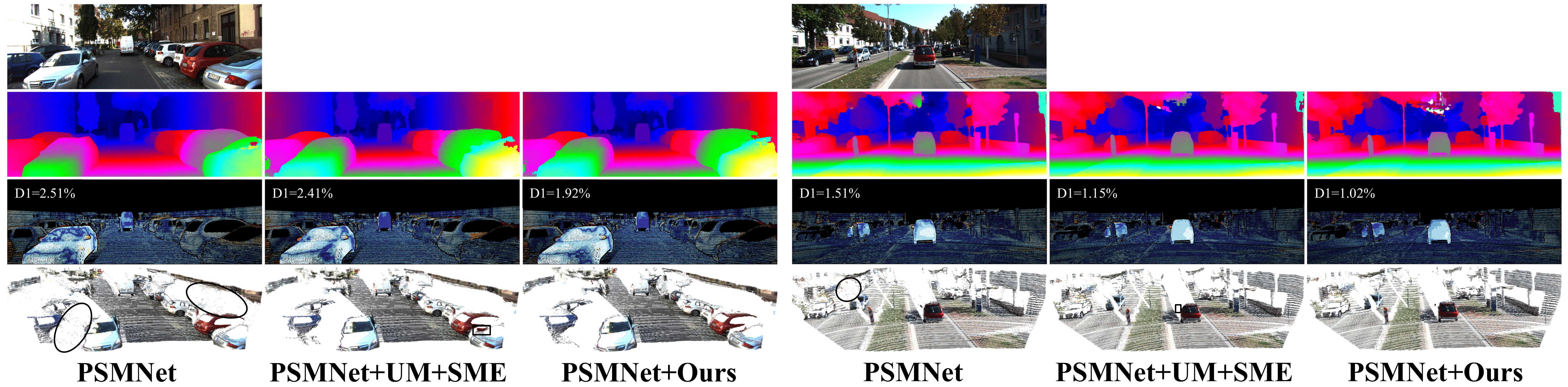}
    \caption{\textbf{Qualitative comparison on KITTI 2015.} From top to bottom: left images, disparity maps, error maps, and reconstructed point clouds. The elliptical and rectangular boxes show partial over-smoothing and misalignment artifacts, respectively.}
    \label{kitti vis} 
\end{figure*}


To have a deeper understanding of the change in the distribution volume brought by our new loss, we count the proportions of pixels with different number of output patterns and list them in~\cref{modal number}. PSMNet~\cite{PSMNet} yields the most uni-modal distributions for all of the pixels, namely, 98.5\%. However, a part of these uni-modal distributions are not correct, as they may be centered on the wrong disparities which lead to large outliers. When uni-modal cross-entropy loss~\cite{SMNet} is employed, the proportion of multi-modal distributions in the edge regions rises from 12.21\% (10.30\%+1.91\%) to 59.84\% (52.70\%+7.14\%), indicating the failure of the uni-modal constraints on the distribution volume. Compared with the uni-modal loss, our adaptive multi-modal loss yields about 5\% more multi-modals at the edge while resulting in lower outliers. This demonstrates the superiority of our loss in supervising the network to produce easily distinguishable modals for disparity estimation. Interestingly, our method achieves better non-edge performance than~\cite{SMNet}, reducing outliers from 2.20\% to 2.07\%. It can also be observed that more uni-modal distributions than~\cite{SMNet} are produced for the non-edge areas, which means that the supervision of clear patterns is beneficial for learning on not only edge but also non-edge pixels.

\cref{probability} shows the output distributions of the two exampling pixels. We can observe that: 1) PSMNet outputs little multi-modal distributions but large disparity error; 2) SME incurs misalignment artifacts from ambiguous distribution; 3) Our loss outputs more easily distinguishable multi-modal distributions than the uni-modal one.

\subsection{Performance evaluation}

We integrate our method into several baseline networks and compare them with other methods.

\textbf{SceneFlow.} As shown in~\cref{SceneFlow Performance Table}, our method significantly improves the performance of all of the baselines by simply changing the loss function and disparity estimator. In particular, the EPE metrics are improved by 41.28\%, 19.48\%, and 35.90\% for the baselines PSMNet, GwcNet, and GANet, respectively. GANet with our method achieves the state-of-the-art results on 1px and 3px metrics. Additionally, our multi-modal trained PSMNet also performs much better than the uni-modal trained one~\cite{SMNet}. Qualitative results shown in~\cref{SceneFlow Qualitative Result} also validate the improvements.

\textbf{KITTI 2015 \& KITTI 2012 benchmarks.}
As the ground-truth of KITTI is sparse~\cite{KITTI2012,KITTI2015}, leveraging the adjacent rows for ground-truth modeling would be beneficial. Therefore, the size of the local window for generating the ground-truth distributions is enlarged to $3\times9$ when fine-tuning the SceneFlow pre-trained network on KITTI datasets. As the results shown in~\cref{KITTI performance}, all of the three baselines are lifted to a highly competitive level by our method. In particular, GANet with our method achieves new state-of-the-art results on both KITTI 2015 and KITTI 2012 benchmarks. Furthermore, we outperform those methods~\cite{PDSNet,SMNet,AcfNet,LaC}, whose loss function contains a uni-modal cross-entropy term, by a large margin.


Since \cite{Non-parametric} doesn't have results on KITTI, we retrain the PSMNet with this method for the purpose of comparison. As shown in~\cref{KITTI performance}, our method also preforms better than those involving multi-modal modeling~\cite{SMDNet,Non-parametric}.

To show our improvements more clearly, we convert the resulting disparity maps to point clouds. As shown in~\cref{kitti vis}, our method dramatically improves the over-smoothing artifacts, and can obtain the point clouds with precise edge structures. More accurate point clouds can be very beneficial for downstream tasks, such as pseudo-LiDAR-based 3D object detection~\cite{pseudo-lidar}.


\subsection{Cross-domain generalization performance}
\label{section generalization}

\begin{table}[t] \footnotesize

\centering
\setlength{\tabcolsep}{6pt}
\begin{tabular}{lcccc}
\toprule
\multirow{2}{*}{Method} & KT 15 & KT 12 &  MB & ETH3D 
 \\
& \textgreater3px & \textgreater3px & \textgreater2px & \textgreater1px \\
\midrule
PSMNet \cite{PSMNet} & 16.3 & 15.1 & 25.1 & 23.8 \\
GwcNet \cite{GwcNet} & 12.8 & 11.7 & 18.1 & 9.0 \\
GANet \cite{GANet} & 11.7 & 10.1 & 20.3 & 14.1 \\
DSMNet \cite{DSMNet} & 6.5 & 6.2 & 13.8 & 6.2 \\
CFNet \cite{CFNet} & 5.8 & 4.7 & 13.5 & 5.8 \\
FC-GANet \cite{FCNet}  & 5.3 & 4.6 & 10.2 & 5.8 \\
Graft-GANet \cite{GraftNet} & 4.9 & 4.2 & 9.8 & 6.2 \\
ITSA-CFNet \cite{ITSA} & 4.7 & 4.2 & 10.4 & 5.1 \\
IGEVStereo \cite{igev} & -- & -- & \textbf{7.1} & 3.6\\
\midrule
PSMNet + Ours & 4.78 & 4.23 & 8.85 & 3.44 \\
GwcNet + Ours & \textbf{4.52} & 4.19 & 9.11 & 3.79 \\
GANet + Ours & 4.84 & \textbf{3.93} & 8.72 & \textbf{2.31} \\
\bottomrule
\end{tabular}
\caption{\textbf{Cross-domain generalization evaluation.}}
\label{cross-domain}
\end{table}

\begin{figure}[t]
    \centering
    \includegraphics[width=0.95\linewidth]{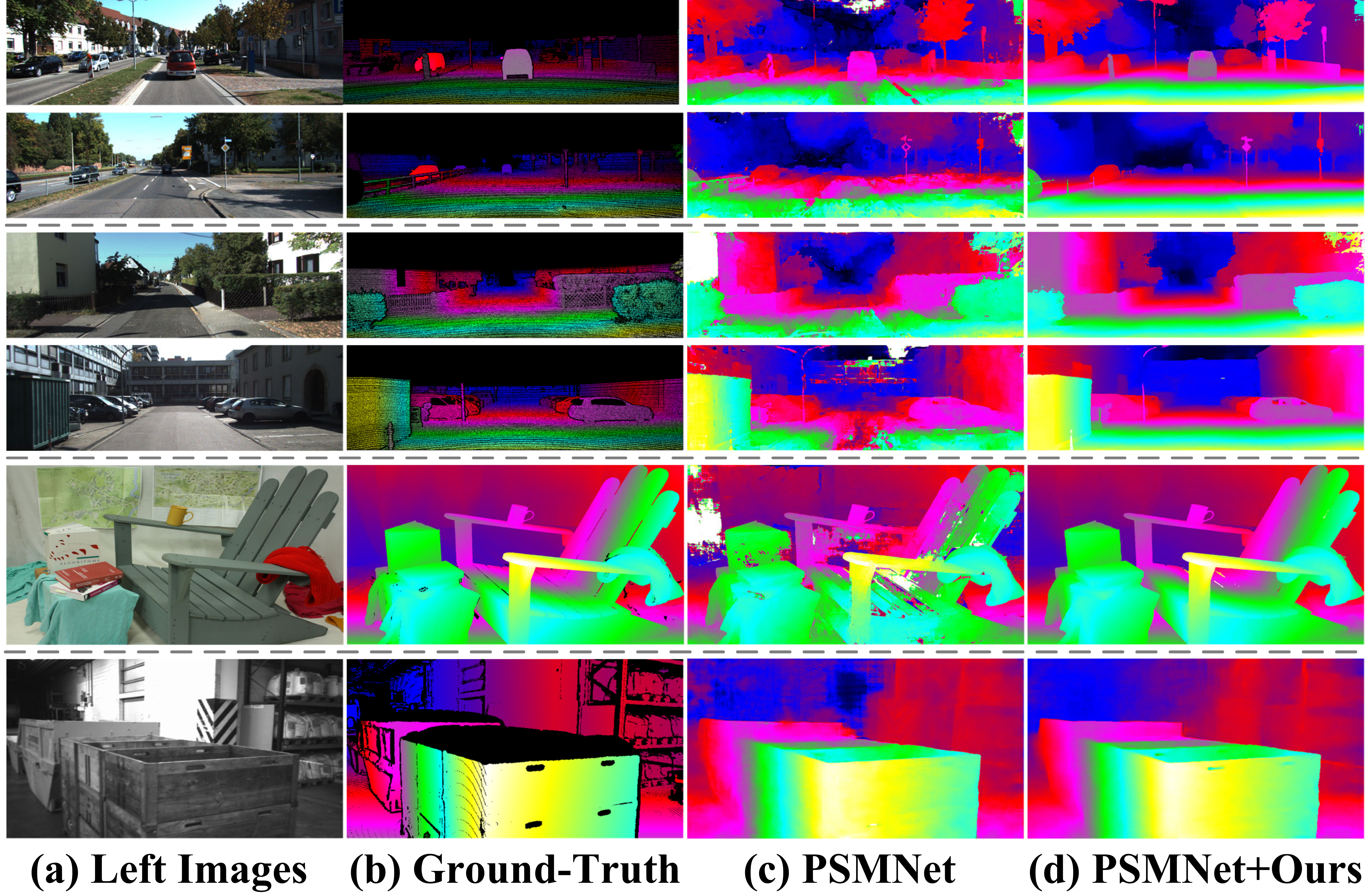}
    \caption{\textbf{Qualitative comparison of cross-domain generalization.} From top to bottom: KITTI 2015 \cite{KITTI2015}, KITTI 2012 \cite{KITTI2012}, Middlebury \cite{Middlebury}, and ETH3D \cite{ETH3D}.}
    \label{generalization vis}
\end{figure}

Besides the fine-tuning performance, generalization is also crucial for deploying networks in the real world. In this section, we compare our generalization performance with baselines, as well as other methods that are specially designed for cross-domain generalization. All methods are only trained on SceneFlow and then tested on four real-world datasets~\cite{KITTI2015,KITTI2012,Middlebury,ETH3D}.

As shown in~\cref{cross-domain}, the generalization performance of the baselines are greatly enhanced by our method. Meanwhile, by guiding the networks to learn explicit multi-modal patterns, our method shows superior performance than existing generalization-focused works on all four datasets. This proves that multi-modal distribution is more in line with the nature of stereo matching.~\cref{generalization vis} shows the qualitative comparison between the original PSMNet~\cite{PSMNet} and ours.

\subsection{Influence of sparser ground-truth}
Acquiring dense and accurate disparity ground-truth are difficult and expensive, especially for outdoor scenes. KITTI 2012 registers consecutive LiDAR point clouds with ICP to increase the ground-truth density~\cite{KITTI2012}, and KITTI 2015 further leverages detailed 3D CAD models to recover points on dynamic objects~\cite{KITTI2015}. Despite these efforts, the valid ground-truth density on KITTI 2015 is only about 30\%. In addition, the error of point cloud registration also needs to be considered, which affects the ground-truth quality. Therefore, a network that can be trained with sparser LiDAR ground-truth will be applauded.

In this experiment, we simulate different densities by randomly down-sampling the original ground-truth on KITTI 2015.~\cref{sparsification} lists the influence of the ground-truth density to the final performance. The performance of both the original PSMNet~\cite{PSMNet} and ours degrades with less supervision signal. However, our method is much less affected than its counterpart, with just 10.13\% degradation when trained with only 20\% of the original ground-truth density. Even with this worst result, it is still much better than the best result of PSMNet trained with 100\% original density. This clearly indicates the large potential of our method in saving the cost of collecting dense ground-truth for real applications.

\begin{table}[t] \footnotesize

\centering
\setlength{\tabcolsep}{12pt}
\begin{tabular}{cll}
\toprule
Density & PSMNet \cite{PSMNet} & PSMNet+Ours\\
\midrule
100\% & 1.90  & 1.58 \\
80\% &  2.15 (-13.16\%) & 1.61 (-1.90\%) \\
60\% &  2.24 (-17.89\%) & 1.68 (-6.33\%) \\
40\% &  2.30 (-21.05\%) & 1.71 (-8.23\%) \\
20\% &  2.35 (-23.68\%) & 1.74 (-10.13\%) \\
\bottomrule
\end{tabular}
\caption{\textbf{Influence of the ground-truth density.} D1 metric is reported on KITTI 2015 validation set.}
\label{sparsification}
\end{table}

\section{Conclusion}
In this work, we propose an adaptive multi-modal cross-entropy loss for stereo matching networks. Contrary to the previous works that impose uni-modal constraints on the distribution volume, our method encourages multi-modal outputs for edge pixels to avoid confusion in network learning. The number of modals and their corresponding weights in the distribution are determined by clustering and statistics within a local window. We also optimize the disparity estimator to robustly locate the dominant modal from the multi-modal outputs. Our method is general and can be easily implemented to enhance the performance of most of the existing stereo networks. GANet with our method achieves the new state-of-the-art on the KITTI 2012 and 2015 benchmarks. Our method is also robust to sparser ground-truth and exhibits excellent cross-domain generalization performance.

\noindent \textbf{Acknowledgement.} We thank all the reviewers for their
valuable comments. The work is supported by the Key Research \& Development Plan of Zhejiang Province (2021C01196).

{
    \small
    \bibliographystyle{ieeenat_fullname}
    \bibliography{main}
}


\end{document}